\documentclass{article}

\usepackage[final, nonatbib]{neurips_2021}

\usepackage[utf8]{inputenc} 
\usepackage[T1]{fontenc}    
\usepackage{hyperref}       
\usepackage{url}            
\usepackage{booktabs}       
\usepackage{amsfonts}       
\usepackage{nicefrac}       
\usepackage{microtype}      
\usepackage{xcolor}         
\usepackage{comment}
\usepackage{graphicx}       
\usepackage{subfig}

\usepackage{biblatex}
\title{DANNTe: a case study of a turbo-machinery sensor virtualization under domain shift}

\author{
  Luca Strazzera \\
  Baker Hughes\\
  \texttt{luca.strazzera@bakerhughes.com} \\
  \and
  Valentina Gori \\
  Baker Hughes \\
  \texttt{valentina.gori@bakerhughes.com} \\
  \and
  Giacomo Veneri \\
  Baker Hughes \\
  \texttt{giacomo.veneri@bakerhughes.com} \\ 
}

\begin{document}

\maketitle

\begin{abstract}
We propose an adversarial learning method to tackle a Domain Adaptation (DA) time series regression task (DANNTe). The regression aims at building a virtual copy of a sensor installed on a gas turbine, to be used in place of the physical sensor which can be missing in certain situations. Our DA approach is to search for a domain-invariant representation of the features. The learner has access to both a labelled source dataset and an unlabelled target dataset (unsupervised DA) and is trained on both, exploiting the minmax game between a task regressor and a domain classifier Neural Networks. Both models share the same feature representation, learnt by a feature extractor. This work is based on the results published by Ganin et al. \cite{ganin2016domain}; indeed, we present an extension suitable to time series applications. We report a significant improvement in regression performance, compared to the baseline model trained on the source domain only.
\end{abstract}

\section{Introduction}
Unsupervised Domain Adaptation \cite{domain_survey}, with generalization bounds stated by Ganin et al. \cite{domain_adaptation_representation}, \cite{domain_adaptation_theorem} is a type of transfer learning \cite{transfer_survey} where the task remains the same while the domains are different (transductive transfer learning). Formally, the learner has access to a labeled source dataset $S = \{(x_i^s, y_i^s)\}_{i=1}^{n_s}$ and an unlabeled target dataset $T = \{(x_i^t)\}_{i=1}^{n_t}$, where $\{x_i^s\}_{i=1}^{n_s}$ follow a probability distribution $P_s$ and $\{x_i^t\}_{i=1}^{n_t}$ follow a target distribution $P_t$. 

We seek to build a domain-invariant feature representation, to ensure good performance on both source and target domain. We apply the unsupervised DA method to an industrial turbo-machinery context providing practical results on a complex timeseries application, even in presence of a non-independently and identically distributed assumption.

\subsection{Related works}
Several lines of research address the unsupervised domain adaptation task. One line aims to learn a 
\textit{domain-invariant feature representation}, typically in the form of a feature extractor neural network \cite{domain_invariant,domain_invariant2,domain_invariant3,domain_invariant4,Li_2020}. Another one aims to learn a \textit{mapping} from one domain to another. The line of \textit{normalization statistics} exploits the batch normalization layer \cite{batch_layer} to learn domain knowledge \cite{domain_batch}. The line of \textit{ensemble methods} consists of using multiple models \cite{domain_ensemble} averaging their output to keep domains separated.

\subsection{Use case}
A turbo-machinery is a system that transfers energy between a rotor and a fluid, including both turbines and compressors. While a turbine transfers energy from a fluid to a rotor, a compressor transfers energy from a rotor to a fluid. 

The turbo-machinery application described in this work consists in building a virtual sensor (ie: a regression model) using data collected from a prototype unit during winter-time and applying it to data collected from the same prototype, during summer-time. The domain shift we are facing is thus mainly related to different ambient conditions, influencing the distribution of the input features.

\subsection{Dataset}

\begin{figure}[h]
    \captionsetup{justification=centering}
    \centering
    {\includegraphics[width=0.32\textwidth]{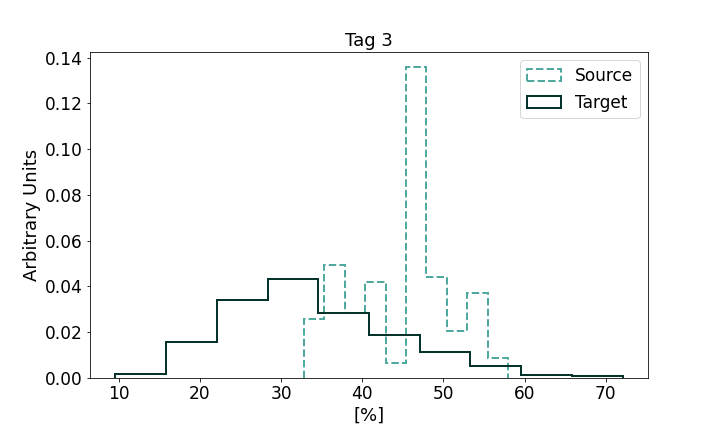}}
    {\includegraphics[width=0.32\textwidth]{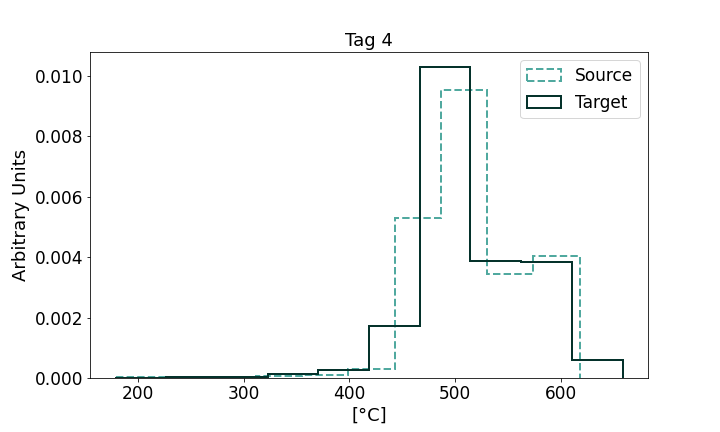}}
    {\includegraphics[width=0.32\textwidth]{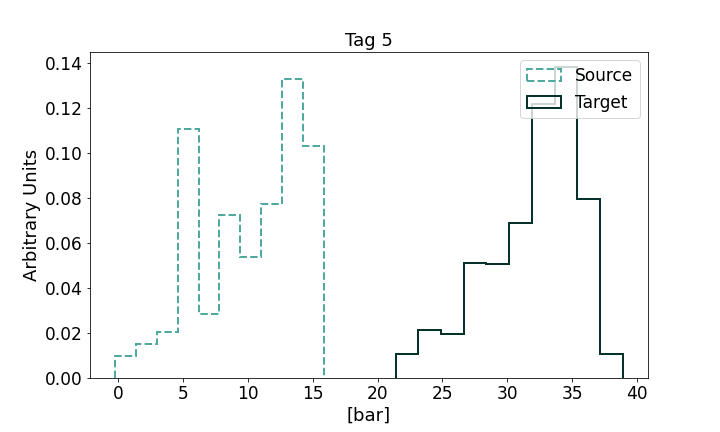}}
    \caption{Distribution of some input features from source (light green) and target (dark green) dataset.}
    \label{fig:feature_dist}
\end{figure}

The dataset \footnotetext{At the moment, dataset cannot be published under open content license due to confidential information.}used to validate the approach is a collection of timeseries acquired from 30 sensors installed on a gas turbine prototype.
Data collected in winter (from December to February) are used as labelled source dataset, while data collected in summer (from June to July) as unlabelled target dataset. 
In Fig.~\ref{fig:feature_dist} the distribution shift (in source and target datasets) of some features inputting the model is shown, as examples.

\section{The implemented library}
\subsection{Model}
We base our solution of the seminal work of Domain-Adversarial Neural Networks (DANN) by Ganin et al. \cite{ganin2016domain}. The idea behind DANN focuses on learning features that combine discriminativeness and domain invariance. This is achieved by jointly optimizing the underlying features as well as two
discriminative classifiers operating on these features. While the
parameters of the classifiers are optimized in order to minimize their error on the training set,
the parameters of the underlying deep feature mapping are optimized in order to minimize the loss of the label classifier and to maximize the loss of the domain classifier. The latter update thus works adversarially to the domain classifier, and it encourages domain-invariant features to emerge in the course of the optimization.

\begin{figure}[h]
\centering
  \includegraphics[scale=0.35]{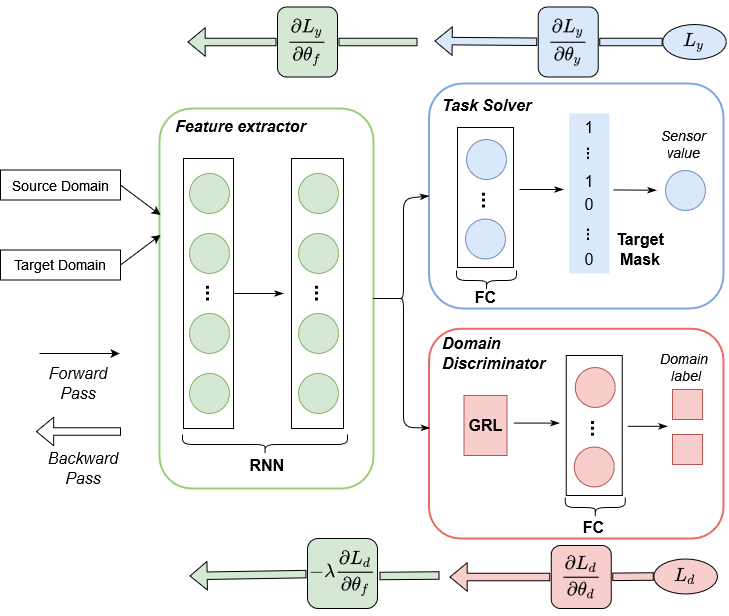}
  \caption{Architecture of our proposed approach (DANNTe), based on Ganin et al. \cite{ganin2016domain}.  Feature extractor weights are modified by both the task solver (in our case, a regressor trying to minimize the reconstruction loss) and the domain classifier (trying to minimize the source vs target domain classification loss). The gradient reversal layer acts so that a minimization problem is solved (instead of a min-max one), just reversing the sign of the domain classifier gradient during backpropagation.}
  \label{fig:dann} 
\end{figure}

Our proposal is based on DANN but focused on a regression task (\textbf{D}omain-\textbf{A}dversarial \textbf{N}eural \textbf{N}etworks applied to \textbf{T}imeseri\textbf{e}s, DANNTe). The architecture is composed of a feature extractor recurrent network feeding a two-headed network (see Fig.~\ref{fig:dann}). The first head is a task solver, in our case a regressor (previously "label predictor") whose goal is to minimize the reconstruction loss for the source domain, where $y$ is available. Its loss is not affected by target domain examples, which are skipped thanks to a target mask layer. The second head is a domain classifier, whose goal is to discriminate examples coming from source from those coming from target domain, exploiting the $x$ information, the only one available in both domains. To train this discriminator, a new dataset $U = \{(x_i, 0)\}_{i=1}^{n_s} \cup \{(x_i, 1)\}_{i=1}^{n_t}$ is built, where 0 and 1 are the domain labels assigned to samples from source domain and target domain respectively.

The task predictor loss $L_y$ is the mean squared error (MSE), while the domain classifier loss $L_d$ is the negative log loss defined as: 
\begin{equation}
    L_d = - \frac{1}{n}\ \sum_{i=1}^n\ [\ y_i\ log(p(y_i)) + (1-y_i)\ log(1-p(y_i))\ ]
\end{equation}

where $y_i$ denotes the domain label for the i-th sample.
\\The total loss then combines the contribution of the losses of the two heads:
\begin{equation}
    L_{tot} = L_y - \lambda L_d  \label{eq:loss}
\end{equation}
where $\lambda$, the domain loss multiplier, influences the contribution of the domain classifier loss during backpropagation.\\ Once trained, only the task solver head (stacked on the feature extractor) is kept and used for inference.

\subsection{Model adaptation}
At training time, Ganin et al.~\cite{ganin2016domain} propose to prepare two datasets: $\{(x_i^s, y_i^s)\}_{i=1}^{n_s}$ to train the \textit{task solver} and  $\{(x_i^s, 0)\}_{i=1}^{n_s} \cup \{(x_i^t, 1)\}_{i=1}^{n_t}$ to train the \textit{domain classifier}. Using two different datasets to train the model, though, causes the need to perform multiple forward and backward passes and thus makes training computationally demanding. To reduce the computational complexity we propose a solution based on the addition of a \textbf{target mask} layer. This layer sets to zero the contribution of the target domain samples to the task solver loss $L_y$. This approach is equivalent to computing the loss $L_y$ first using only samples from the source domain, and then computing the loss $L_d$ using the combined domain batches.

Another proposal by Ganin et al.~\cite{ganin2016domain} is to build the datasets by i.i.d from $P_s$ and $P_t$. In our specific use case, though, the i.i.d. assumption does not hold since we are dealing with timeseries and we use an RNN (LSTM) as a \textit{feature extractor}, so we do not want to lose the time correlation in our dataset. The solution we propose is to train the model by creating \textbf{equally divided batches} where half of each batch is filled with samples from the source domain, and half with samples from the target domain, maintaining the time ordering.

\section{Results}
\subsection{Performance assessment strategy}
Model performance has been evaluated by exploiting the variable $y$ which is available in both domains, in this simplified use case. We remind once again that this ground truth will not be available, instead, in our future and more realistic use case.

The DANNTe regression performance has been compared to:
\begin{itemize}
\item the \textit{constant mean} regressor, where output is always the mean of the source data
\item the \textit{baseline} model to estimate the lower-bound performance, given by a supervised model trained only using source domain data and then directly applied to target domain data.
\item the \textit{fully-supervised} model to estimate the upper-bound performance, given by a supervised model where both source and target domains are labelled and available
\item the original \textit{DANN} method that doesn't take into account the temporal structure of the data, to see if accounting for temporal correlations is helping.
\end{itemize}

\subsection{Model performance and comparison}
Performance comparison summarized in Tab.~\ref{tab:results} shows the results of our experiments with related uncertainties that refer to the variance obtained with k-fold cross-validation. Mean Squared Errors are computed with the y scaled to have zero mean and unit variance. MAPE is computed without scaling the y variable. The last column, where is present the KL-divergence, is computed with the embedding of the two distributions (Source and Target) obtained from the second last layer of each model.   
\begin{table}[h]
  \centering
  \begin{tabular}{lrrrr}
    Model & MSE (Source) & MSE (Target) & MAPE (Target)  & KL\textit{-divergence} \\
    \midrule
    Constant (mean) & 99.9 \scriptsize{$\pm$ 0.0} & 77.5 \scriptsize{$\pm$ 0.0} & 12.0 & - \\ 
    Baseline & \textbf{0.7} \scriptsize{$\pm$ 0.1} & 5.1 \scriptsize{$\pm$ 0.9} & 3.4 & 6.1\\
    Fully-supervised & 0.9 \scriptsize{$\pm$ 0.7} & \textbf{1.8} \scriptsize{$\pm$ 0.8} & \textbf{0.9} & \textbf{1.2} \\ 
    \midrule
    DANN & 0.8 \scriptsize{$\pm$ 0.3} & 3.6 \scriptsize{$\pm$ 0.6} & 2.5 & 2.7 \\ 
    \textbf{DANNTe} & 0.8 \scriptsize{$\pm$ 0.4} & 2.3 \scriptsize{$\pm$ 0.4} & 1.5 & 2.5\\ [0.05cm]
    \hline
    \hline
    \vspace{\baselineskip}
  \end{tabular}
  \vspace{\baselineskip}
  \caption{DANNTe performance compared to baseline and fully-supervised models. MSE (Source) and MSE (Target) are both in the $10^{-2}$ format. MAPE (Target) is in the $10^{-1}$ format. KL\textit{-divergence} is in the $10^{-1}$ format.}
  \label{tab:results}
\end{table}

DANNte performs worse on source domain (MSE Source) compared to the beseline model, but improves performance in the target domain (MSE Target and MAPE Target) with respect to baseline and DANN models. This is mainly due to its ability to manage temporal dependencies. 

We found that the hyperparameter $\lambda$ (see Eq.~\ref{eq:loss}) plays a key role in feature extraction. The larger its value, the higher is the domain classifier importance in the feature extractor training.
In our experiments, for our use case we found that $\lambda=1.5$ yields the best performance.

\subsection{Features encoding}
\begin{figure}[h]
    \captionsetup{justification=centering}
    \centering
    \subfloat[Baseline]{\includegraphics[width=0.30\textwidth]{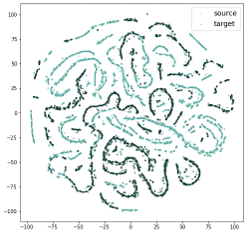}\label{fig:worst_case_umap}}
    \subfloat[Fully supervised]{\includegraphics[width=0.30\textwidth]{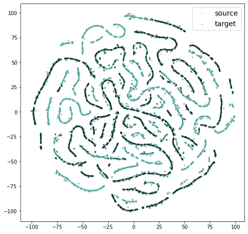}\label{fig:best_case_umap}}
    \subfloat[DANNte]{\includegraphics[width=0.30\textwidth]{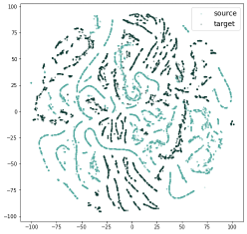}\label{fig:DARN_umap}}
    \caption{Two-dimensional data representation using UMAP. We can observe a clear separation between the two domains (light green and dark green, source and target respectively) using the baseline model (a). Both Fully Supervised (b) and DANNTe (c) construct a less discriminative feature representation.}
    \label{fig:umaps}
\end{figure}

To get a graphic insight about how the feature embedding changes with DANNTe, we use the UMAP method \cite{umap} to visualize it (see Fig.~\ref{fig:umaps}). 
In the fully supervised model (Fig.~\ref{fig:best_case_umap}) a scattered separation between source and target domains can be seen with respect to the baseline model (Fig.~\ref{fig:worst_case_umap}), verified by the KL divergence in Tab.~\ref{tab:results}.
Moreover, the proposed DANNTe approach shows an higher ability to superimpose different domains (Fig.~\ref{fig:DARN_umap}) with respect the baseline model.

\section{Conclusions and Future work}
We have shown a domain adaptation method applied to a regression task for an industrial use case.
The described technique allows a signal virtualization based on embedded features that combine good performance on task and domain invariance.
The results are promising and allow us to improve the reliability of a model on a target domain. With respect to a baseline model trained using only data sampled from a source domain and the DANN model, it achieves better performance on the target domain. Despite there is still room for improvement to obtain results as close as possible to the fully-supervised approach, our approach brings valuable results which can be relied upon in real applications. 
Our future use case will develop a virtual sensor from a prototype unit and applying it to a fleet unit installed at Customer site. Therefore, domain shift will be due not only to different ambient conditions but also to different operative conditions of the machines.

\section*{Acknowledgement}
Special thanks to Andrea Panizza, Giacomo Monaci, Marzia Sepe and Valeria Ballarini for the valuable discussion.

\printbibliography

\end{document}